\documentclass{article}

\usepackage{microtype}
\usepackage{graphicx}
\usepackage{subfigure}
\usepackage{booktabs} 

\usepackage{hyperref}

\usepackage{listings}

\usepackage[accepted]{icml2025}

\usepackage{amsmath}
\usepackage{amssymb}
\usepackage{mathtools}
\usepackage{amsthm}
\usepackage{comment}

\usepackage[capitalize,noabbrev]{cleveref}

\theoremstyle{plain}

\theoremstyle{definition}

\theoremstyle{remark}

\usepackage[textsize=tiny]{todonotes}

\begin{document}

\twocolumn[
\icmltitle{Fooling the Watchers: Breaking AIGC Detectors via Semantic Prompt Attacks}



\icmlsetsymbol{equal}{*}

\begin{icmlauthorlist}
\icmlauthor{Run Hao}{Aarhus}
\icmlauthor{Peng Ying}{cumt_1,cumt_2}
\end{icmlauthorlist}
\icmlaffiliation{Aarhus}{Aarhus University}
\icmlaffiliation{cumt_1}{School of Computer Science and Technology, China University of Mining and Technology, Xuzhou, China}
\icmlaffiliation{cumt_2}{Mine Digitization Engineering Research Center of the Ministry of Education, China University of Mining and Technology, Xuzhou, China}


\icmlcorrespondingauthor{Run Hao}{au785440@uni.au.dk}
\icmlcorrespondingauthor{Peng Ying}{pengying@cumt.edu.cn}

\icmlkeywords{Adeveserial Attack}

\vskip 0.3in
]



\printAffiliationsAndNotice{\icmlEqualContribution} 

\begin{abstract}
The rise of text-to-image (T2I) models has enabled the synthesis of photorealistic human portraits, raising serious concerns about identity misuse and the robustness of AIGC detectors. In this work, we propose an automated adversarial prompt generation framework that leverages a grammar tree structure and a variant of the Monte Carlo tree search algorithm to systematically explore the semantic prompt space. Our method generates diverse, controllable prompts that consistently evade both open-source and commercial AIGC detectors. Extensive experiments across multiple T2I models validate its effectiveness, and the approach ranked first in a real-world adversarial AIGC detection competition. Beyond attack scenarios, our method can also be used to construct high-quality adversarial datasets, providing valuable resources for training and evaluating more robust AIGC detection and defense systems.
\end{abstract}

\section{Introduction}








In recent years, text-to-image (T2I) models have witnessed significant advancements, driven by breakthroughs in deep learning and generative modeling. Early works such as AttnGAN~\cite{xu2018attngan} and StackGAN~\cite{zhang2017stackgan} laid the foundation for generating coherent images from textual descriptions by integrating attention mechanisms and multi-stage refinement strategies. With the emergence of large-scale diffusion models, notably DALL$\cdot$E~2~\cite{ramesh2022hierarchical}, Imagen~\cite{saharia2022photorealistic}, and Stable Diffusion~\cite{rombach2022high}, the quality and realism of generated images have drastically improved, enabling high-fidelity synthesis conditioned on natural language prompts.

Alongside the rapid development of T2I models, the proliferation of AI-generated content (AIGC) has raised concerns regarding authenticity and misuse. In particular, their ability to synthesize photorealistic human portraits has introduced new challenges in identity and privacy protection. Hyper-realistic faces can be fabricated with arbitrary attributes, potentially enabling malicious actors to create fake IDs, impersonate real individuals, or even generate sensitive visuals of non-consenting subjects~\cite{karras2019style, chesney2022deep, douglas2021deep}. Such misuse raises pressing ethical and legal questions, particularly when these synthetic faces are indistinguishable from real ones by both humans and machines~\cite{mirsky2021creation, verdoliva2020media}.
This has led to growing interest in AIGC detection, which aims to distinguish between real and synthetic content. Recent works in this domain, such as GAN-generated image detectors~\cite{wang2020cnn} and diffusion-specific classifiers~\cite{wang2023dire}, have leveraged both low-level artifacts (e.g., color statistics~\cite{durall2020watch}, sensor noise~\cite{frank2020leveraging}, frequency domain inconsistencies~\cite{durall2020watch}) and high-level semantic cues (e.g., unnatural textures~\cite{zhang2019detecting}, unrealistic object layouts, or prompt-image mismatches~\cite{chen2023prompt, liang2023mindeye}) to achieve robust detection.

Despite the increasing sophistication of these detectors, many rely on assumptions about known generation models or artifacts, leaving them vulnerable to adaptive or black-box attacks. In high-stakes scenarios involving human portraits, failure to detect synthetic content may lead to severe consequences, including misinformation, reputational harm, or privacy breaches, motivating the need for rigorous evaluation of detector robustness under adversarial conditions.
However, conducting such evaluations poses several unique challenges in the context of T2I generation. \textit{First, the inherently stochastic nature of modern T2I models limits the precision with which image content can be controlled.} Unlike traditional adversarial attacks that directly manipulate pixel-level features or inject specific noise patterns, prompt-based attacks must rely on indirect, semantic-level perturbations—often resulting in only approximate control over the generated outputs~\cite{nichol2021glide}. \textit{Second, the process of manually crafting effective prompts is prohibitively costly and inefficient.} Identifying combinations of textual cues that consistently induce detection failures requires substantial human effort and insight, and it remains difficult to determine which elements of a prompt are most responsible for the success or failure of an attack.

To overcome these challenges, we develop a grammar tree-based prompt generation mechanism, where each node in the tree corresponds to a semantic component (e.g., portrait attributes or stylistic features), and prompts are constructed through a recursive top-down traversal of this tree. To further optimize this process, we integrate a variant of the Monte Carlo Tree Search (MCTS) algorithm—specifically, the UCT-Rand algorithm~\cite{zheng2024reqsminer}—which replaces greedy selection with weighted random sampling. This allows for broader and more effective exploration of the prompt space, increasing the likelihood of generating photorealistic portraits capable of evading both open and closed-source AIGC detectors. To demonstrate the practical effectiveness of our approach, we deployed the proposed method in a real-world AIGC adversarial detection competition~\cite{tencent_zhuque_2025}, where it achieved first place. Furthermore, we applied our method across multiple state-of-the-art T2I models and evaluated its performance against both open-source and commercial AIGC detectors. Experimental results show that existing detectors are highly susceptible to prompts generated by our framework, further highlighting the fragility of current detection mechanisms when facing prompt-level semantic attacks. 

Our contributions are summarized as follows:
\begin{itemize}
\item {We propose semantic prompt attack methods targeting both closed-source and open-source AIGC detectors, revealing detector's susceptibility to such attacks. To the best of our knowledge, this is the first  study on prompt-level semantic attacks against AIGC detectors.}

\item{We design a hierarchical grammar tree to structure semantic segments of prompts, such as portrait attributes and stylistic features to enable automated and controllable prompt generation, thereby allowing text-to-image models to produce images that can bypass AIGC detectors.}


\item{ Our method achieves strong evasion performance across both open- and closed-source detectors, and ranks first in a real-world competition. Furthermore, the generated prompts can serve as valuable adversarial datasets to support the development and evaluation of more robust and adaptive detection models.}
\end{itemize}

\section{Related Work}
\subsection{AI-generated Image Detection}

Recent advancements in AIGC detection have introduced a range of effective defense strategies. Zhong et al.~\cite{zhong2024patchcraftexploringtexturepatch} propose a method that suppresses global semantic information and enhances local texture features by analyzing inter-pixel correlations between rich and poor texture regions, improving generalization across unseen generative models. Frank et al.~\cite{frank2020leveraging} reveal that GAN-generated images exhibit consistent artifacts in the frequency domain caused by upsampling operations, which can be effectively exploited for detection. Liu et al.~\cite{liu2022detecting} focus on learning the noise patterns inherent to real images in both spatial and frequency domains, enabling the detection of generated images by identifying deviations from these natural patterns. Zhengzhe Liu et al.~\cite{liu2020global} design a detector that extracts global texture representations using Gram matrices, achieving robustness to image distortions and strong generalization to unseen fake images. Cazenavette et al.~\cite{cazenavette2024fakeinversion} extract inversion features from a pre-trained diffusion model by reconstructing latent noise maps and denoised outputs, allowing their detector to generalize well to high-fidelity text-to-image models. Luo et al.~\cite{luo2024lare} propose a lightweight approach that calculates latent-space reconstruction error in a single denoising step and refines image features based on spatial correlations with this error, improving both detection accuracy and efficiency.

\subsection{Datasets for AI-generated Image Detection}


CNNSpot\cite{wang2020cnngeneratedimagessurprisinglyeasy} collects images generated by 11 CNN-based models (e.g., ProGAN\cite{kang2023scalingganstexttoimagesynthesis}) to build a dataset that covers diverse architectures and tasks, enabling evaluation of the detector’s cross-generator generalization ability.
GenImage\cite{zhu2023genimage} adopts a unified prompt template \verb|"photo of {class}"|, where \verb|{class}| represents one of the 1000 categories from ImageNet\cite{5206848}, to generate class-controlled images.
WildFake\cite{hong2024wildfakelargescalechallengingdataset} collects real user-generated prompts from community platforms such as Civitai\cite{civitai2025} and Midjourney\cite{holub2022midjourney}, and supplements them with its own generation pipeline. The prompts cover various styles, including stylized, personalized (e.g., DreamBooth\cite{ruiz2022dreambooth}, LoRA\cite{hu2021loralowrankadaptationlarge}), and structure-controlled (e.g., ControlNet\cite{zhang2023adding}) types.  While prior work\cite{10156981} has investigated prompt-based generation of realistic faces using Stable Diffusion, its reliance on human evaluation limits the possibility of conducting systematic, model-based assessments of detection performance.


\subsection{Black-box Adversarial Attack Methods}
Black-box adversarial attack means that the attacker can not access the internal structure of the model and can only optimize the attack method by querying the model's output. Some methods based on random pixel updates \cite{vo2024brusleattackqueryefficientscorebasedblackbox,andriushchenko2020squareattackqueryefficientblackbox} are effective in evading detection.
Another attack method\cite{meng2023avainconspicuousattributevariationbased} perturbs the semantics in the attribute space of  images to deceive detection models. Prior work\cite{xie2024fakerealrealisticlikerobust} uses real-world post-processing, i.e.,Gaussian blur, JPEG compression, Gaussian noise and light spot to generate adversarial examples. 





\section{Threat Model \& Motivation}
\label{sec:motivation}

\subsection{Threat Model}

We consider an adversary who aims to generate  portrait images capable of bypassing AIGC detectors by crafting carefully designed text prompts for T2I models such as Flux, Midjourney, and Stable Diffusion. We focus exclusively on images generated directly by the T2I model, meaning that any form of post-processing is out of scope. The adversary has black-box access to the AIGC detectors, they can submit images and obtain detection scores, but do not know the detector’s  architecture.

The generated images must satisfy several predefined constraints:(1) facial features must be proportionally accurate; (2) expressions should appear natural; (3) the image should conform to basic principles of physical plausibility; (4) scenes must align with common real-world experiences and social norms.

\begin{figure}[tbp]
\vskip 0.1in
\begin{center}
\centerline{\includegraphics[width=\columnwidth]{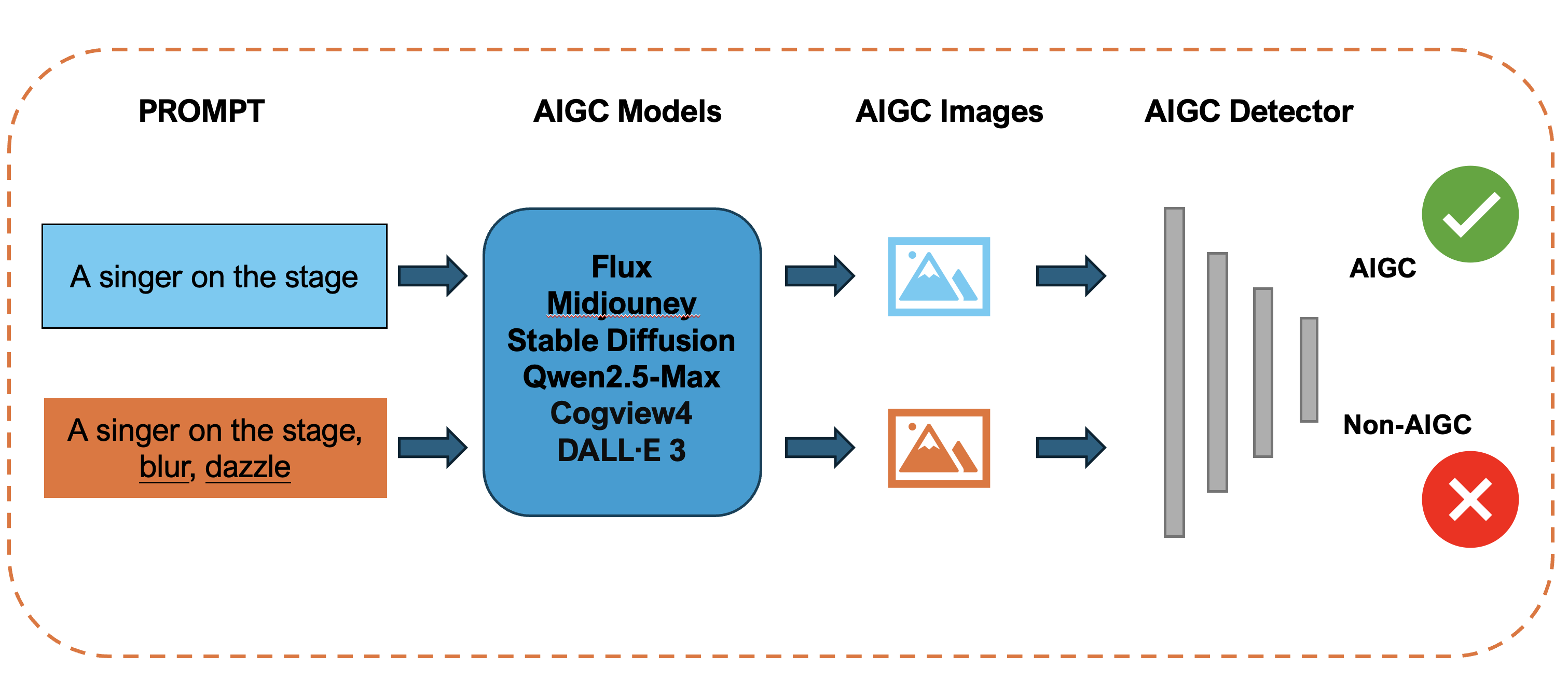}}
\caption{The adversary uses a lighting-based attack to generate images that can evade the detector.}

\label{threatmodel}
\end{center}
\vskip -0.2in
\end{figure}

As shown in Figure \ref{threatmodel}, a simple prompt such as \textit{“A singer on the stage”} generates an image that is detectable as AI-generated. However, by adding a simple constrain “blur, dazzle” in the prompt, the adversary can manipulate the image's brightness, leading it to be classified as a non-AIGC image.







\subsection{ Prompt-based Texture Injection Attack}
PatchCraft\cite{zhong2024patchcraftexploringtexturepatch} is a recently proposed framework for detecting AI-generated images. The core of PatchCraft lies in the assumption that although some cutting-edge generative models create impressive images from the semantic view, it is still hard to simulate the inter-pixel correlation of real images. Pixels in rich texture regions exhibit more significant fluctuations than those in poor texture regions, so it proposes a Smash\&Reconstruction pre-processing method to extract rich and poor texture regions from the image independently, leveraging the inter-pixel correlation contrast between rich and poor texture regions of an image as a fingerprint to identify AI-generated images.

While this method leads to strong generalization across unseen generators, it also exposes a critical vulnerability: if an attacker can manipulate the distribution or content of rich-texture regions, the detector will focus only on the injected rich-texture regions designed by the attacker, making it easy to mislead the detector.

To exploit this weakness, we design an adversarial attack against PatchCraft detector by injecting carefully crafted rich-texture regions into the generated image via prompts. Specifically, we use prompts of the form:
\begin{quote}
    \texttt{<A description of a person> + "with the rest of the page filled entirely with clear text."}
\end{quote}

\begin{figure}[tbp]
\vskip 0.2in
\begin{center}
\centerline{\includegraphics[width=\columnwidth]{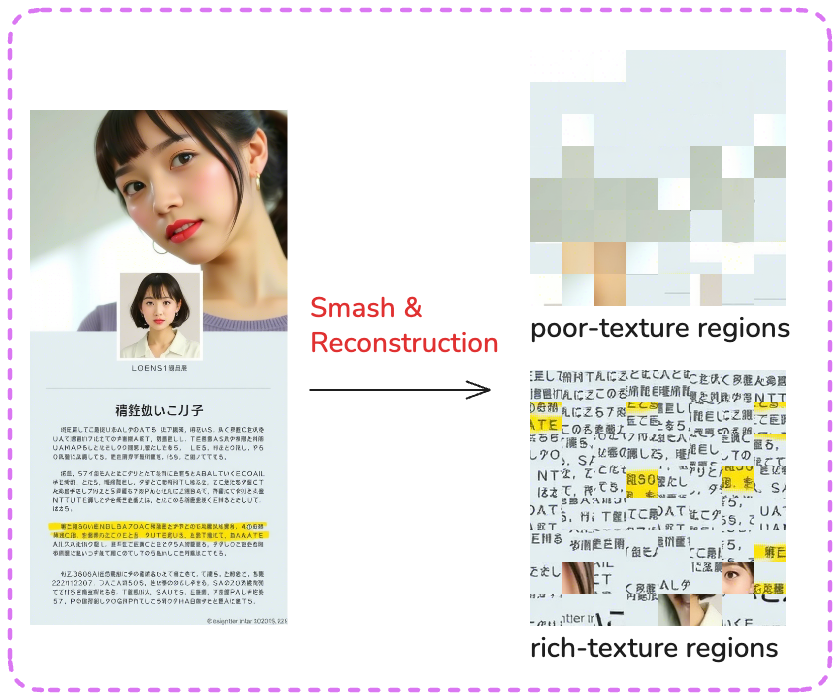}}
\caption{Generated by the flux-schnell model using the prompt: \texttt{<A portrait of a person>} + "with the rest of the page filled entirely with clear text."}

\label{attackfigure1}
\end{center}
\vskip -0.2in
\end{figure}

The principle of the attack is shown in \cref{attackfigure1}, the description prompt ensures that the T2I model still produces a facial region as the primary content, while the ``with the rest of the page filled entirely with clear text'' segment guides the model to fill rest areas with densely structured textual content. 
Through simple testing, we found that AI-generated images composed of clear text and blank space are difficult for AIGC detectors to identify as AI-generated. These textual regions exhibit abrupt pixel variations in the image space, characteristics typical of high-frequency signals, which can be extracted by PatchCraft as rich-texture regions.

By injecting enough patches of artificial text, we effectively dominate the reconstructed rich-texture regions with adversarial content. According to the original intention of the method, the facial regions of a portrait should be extracted as rich-texture regions, but after the Smash \& Reconstruction process, the detector focuses  on the injected texture regions rather than the meaningful fingerprint from the face regions. Meanwhile, the poor-texture regions mostly consist of blank or low-information areas, further limiting the ability of detector to make reliable decisions.

Essentially, our attack disrupts the fundamental texture distribution that PatchCraft relies on. The classifier is misled to learn from manipulated features originating from adversarial prompts rather than natural generation artifacts, thereby inducing systematic detection failures.

\subsection{Prompt-based lighting attack}
\label{subsec:cs2}


In recent years, researchers have increasingly recognized that visual variations caused by lighting in natural images can serve as effective adversarial perturbations to mislead deep neural networks~\cite{tian2021avaadversarialvignettingattack,jiang2023evadingwatermarkbaseddetection,xie2024fakerealrealisticlikerobust}. Unlike traditional adversarial examples that often rely on high-frequency noise, such light-based perturbations are typically more perceptually natural and harder to detect.
Light-based perturbations have emerged as a natural and imperceptible means to evade deep learning models. 

Early work~\cite{tian2021avaadversarialvignettingattack} introduced adversarial vignetting attacks, simulating radial brightness falloff based on physical camera parameters. 
Subsequent research~\cite{jiang2023evadingwatermarkbaseddetection}  revealed that AIGC detectors are vulnerable to global brightness and contrast shifts. Even simple linear transformations were sufficient to mislead models, indicating a lack of robustness to natural lighting variations.
Building on this, R2BA~\cite{xie2024fakerealrealisticlikerobust} proposed using localized light spots to simulate overexposure effects. 


Since we are unable to apply direct post-processing to generate images in our attack scenario, we explore an alternative approach by manipulating lighting conditions through prompt engineering. Specifically, we aim to induce light-based perturbations such as dazzle or overexposure. By incorporating semantically relevant terms (e.g., \textit{“dazzle”}) into the text prompts fed to the T2I model, this action will substantially alter the intensity distribution of pixels within the image, thereby influencing the detector's decision-making process.

Inspired by prior work on adversarial vignetting and light spot attacks, which demonstrate that changes in lighting can impair detector performance, we aim to replicate similar effects at the T2I model generation stage using text prompts. To assess this, we evaluate the impact of prompt-induced lighting perturbations on the Zhuque AIGC detector. Our experiments reveal that incorporating light-related words into prompts (e.g., \textit{“dazzle”})  can significantly reduce detection accuracy, indicating that the detector is sensitive to such light-related semantic cues. 

\begin{figure}[tbp]
\vskip 0.2in
\begin{center}
\centerline{\includegraphics[width=0.3\columnwidth]{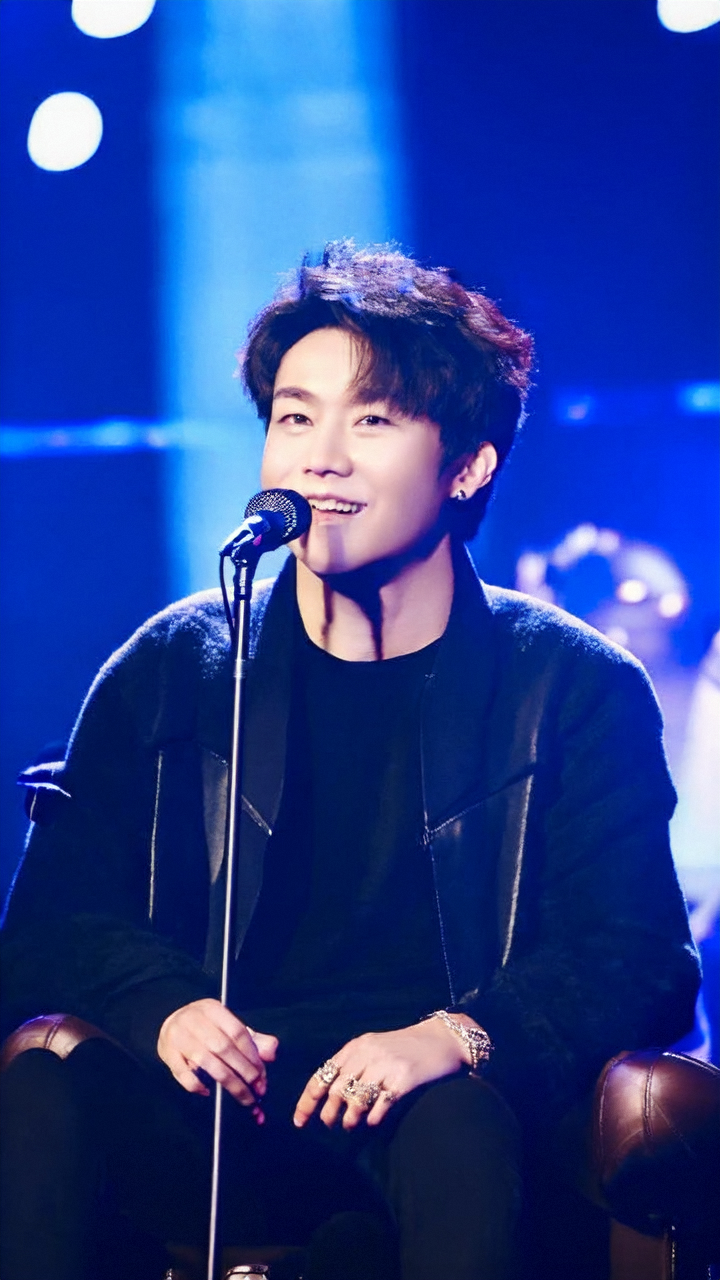}}
\caption{Generated using the wanx2.0-t2i-turbo model with the prompt: 'Jay Chou's live concert, clear facial features, dazzle.' The Zhuque AIGC Detector estimates a 24.3\% probability that the image is AI-generated.
}
\label{attackfigure2}
\end{center}
\vskip -0.2in
\end{figure}

\cref{attackfigure2}  presents a  example in which we add the word “dazzle” to the prompt to control the image's lighting. This manipulation successfully induces a perturbation attack against the AIGC detector.

\section{Method}

\subsection{Overview}



As discussed in the case study in  Section \ref{sec:motivation}, T2I models can generate images that perturb AIGC detectors by controlling image semantic features such as texture and lighting. However, manually constructing prompts to generate images is both costly and inefficient, and it is difficult to find the optimal prompt.

To address this challenge, we developed a grammar tree-based automated prompt generator. 
The root node of the grammar tree is the prompt, and under the root node are several child nodes, such as "Portrait attributes" and "Stylistic attributes". The prompt generator traverses the grammar tree recursively from top to bottom to generate the prompt. We will introduce the design of the grammar tree in detail in Section 4.3.

To enable the prompt generator to efficiently create prompts that can guide T2I models to evade AIGC detectors, we utilized a variation of the Monte Carlo Tree Search (MCTS) algorithm, called the UCT-RAND algorithm~\cite{zheng2024reqsminer}.
UCT-Rand algorithm uses weighted random selection rather than the argmax function to choose the next child node which allows us to comprehensively explore different branches of the grammar tree.

\begin{figure}[tbp]
\vskip 0.2in
\begin{center}
\centerline{\includegraphics[width=\columnwidth]{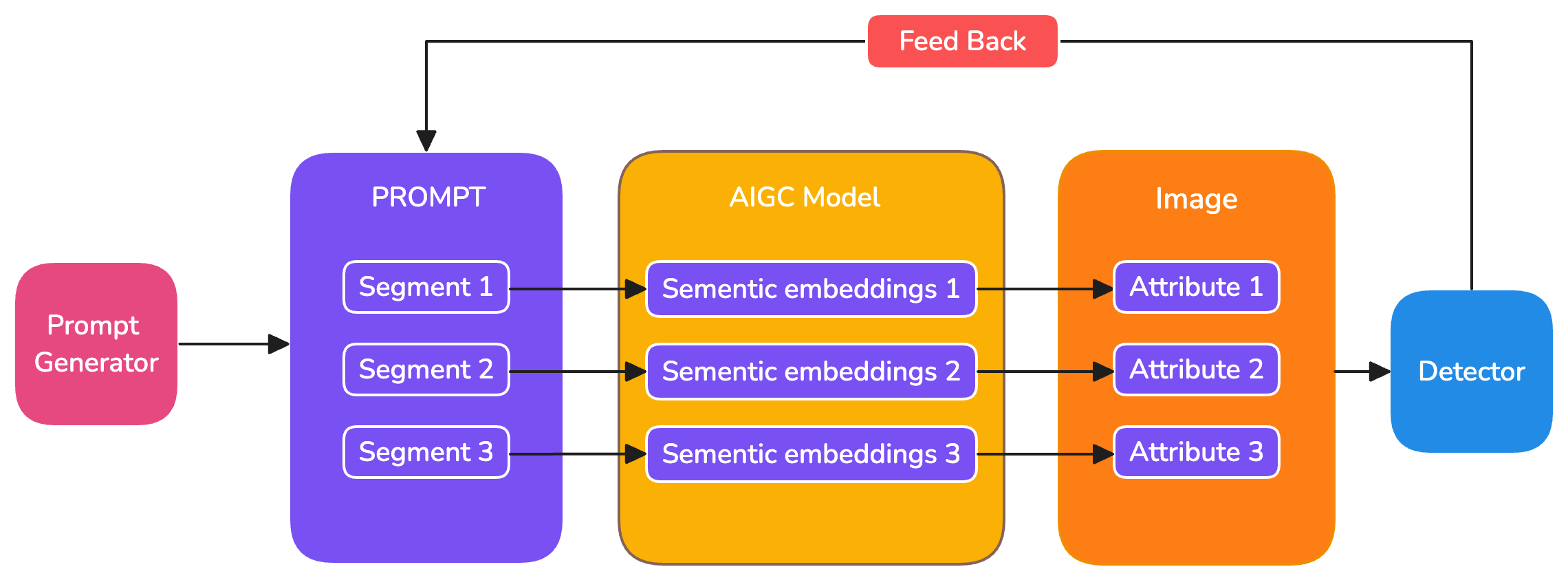}}
\caption{A illustration for generating prompts designed to bypass the AIGC detector.}

\label{featuregraph}
\end{center}
\vskip -0.2in
\end{figure}


\subsection{Definition of Attribute and Semantic Embeddings}

Previous research~\cite{meng2023avainconspicuousattributevariationbased} on attacking DeepFake detectors involved introducing attribute-level perturbations to change the semantics of attributes such as mouth, hairstyle, and eyebrows. In our work, we also use attributes to represent the semantic features in the images,  and we leverage semantic embeddings to represent these attributes within the T2I model's embedding space.

In this paper, we consider two types of attributes: portrait attributes and stylistic attributes. \textbf{Portrait attributes} represent human-interpretable portrait semantic features such as gender, age, and skin tone.
\textbf{Stylistic attributes} include properties such as lighting and texture. For example, a T2I model can generate a image with specific lighting (the attribute), such as "overexposure" (the semantics).

The attack process can be summarized as: carefully crafted segments within the prompt can further influence the semantic embeddings in the T2I model, thereby achieving control over the attributes of the generated image, as shown in~\cref{featuregraph}.

\subsection{Grammar Tree Design}

\begin{figure}[tbp]
\vskip 0.2in
\begin{center}
\centerline{\includegraphics[width=\columnwidth]{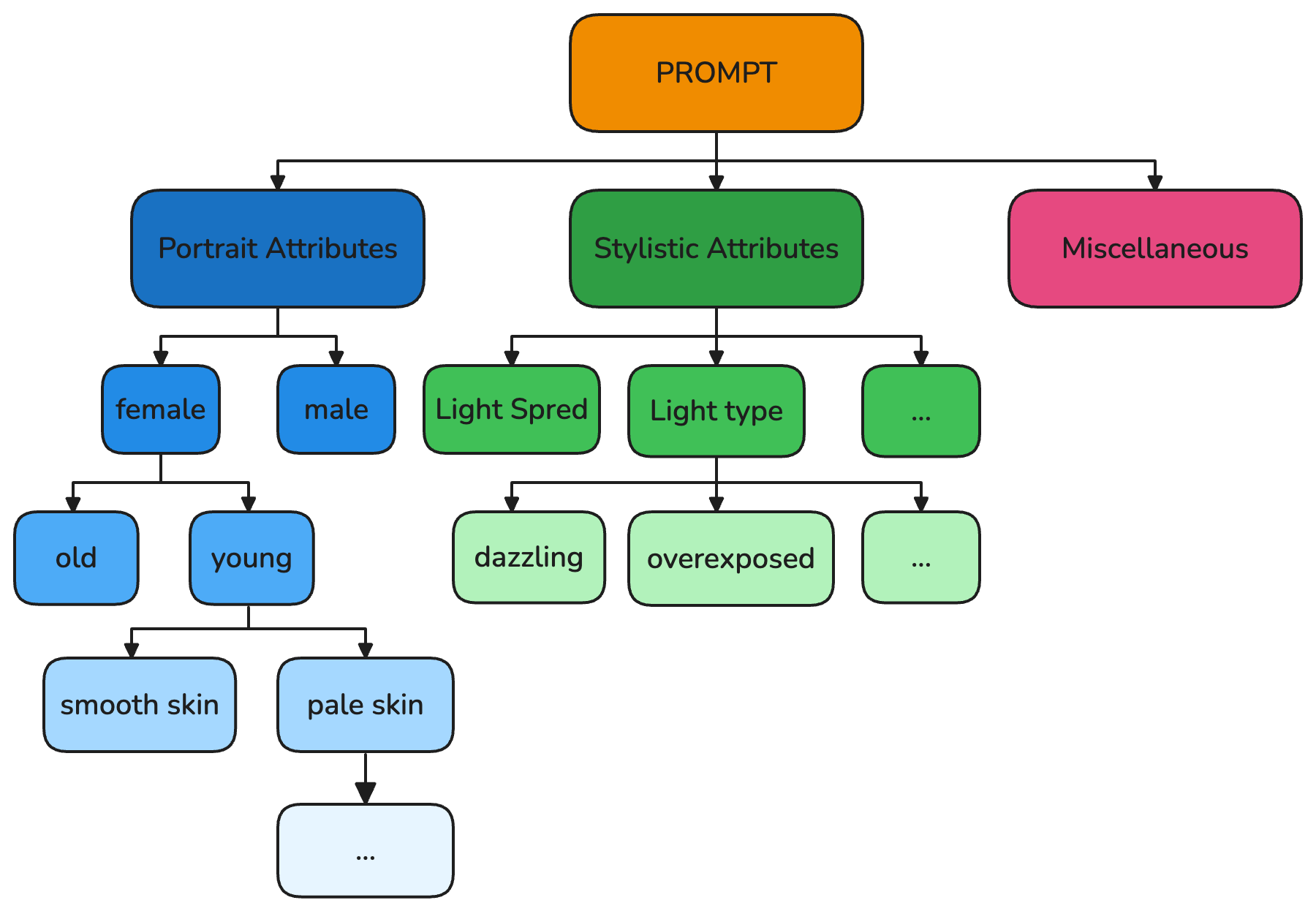}}
\caption{Illustration of the Grammar Tree.}

\label{tree}
\end{center}
\vskip -0.2in
\end{figure}

We believe that an image's attributes are hierarchical, so when designing the grammar tree, we categorize the prompt into three components as shown in~\cref{tree}: portrait attributes, stylistic attributes, and miscellaneous attributes. Portrait attributes are primarily used to control portrait-related semantic features such as age, gender, and skin tone. Stylistic attributes focus on semantic features like lighting and texture. The miscellaneous node includes additional constraints, for instance, "well-defined facial features" ensures the generated images have clear facial structures.

There are three node types in our designed grammar tree: RAND, OR, and AND. The AND node means that all child nodes must be visited and combined in sequence. The OR node indicates a selection among multiple candidate child nodes, where only one is chosen. The RAND node means that the number of times its child node is visited is random. If an attribute contains semantically conflicting features, we represent them using an OR node. If the features are semantically independent, we model them with an AND node. For example, the relationship between Portrait Attributes and Stylistic Attributes. When a specific semantic feature of an attribute may occur once or multiple times, we use a RAND node.

\subsection{Grammar-based Prompt Generator}



Prompt Generator is designed to generate high-quality prompts from a grammar tree to guide AIGC model in producing images that can evade content detectors. In this process, the Prompt Generator treats a designated non-terminal symbol (e.g., \verb|PROMPT|) as the root node and recursively traverses the grammar tree from top to bottom. The leaf nodes serve as the terminal symbols in generated prompt.

To achieve both the efficient generation of adversarial prompts and sufficient exploration of the grammar space, we adopt the  UCT-RAND algorithm~\cite{zheng2024reqsminer}. The generation procedure comprises four phases: Expansion, Selection, Simulation, and Backpropagation.


\textbf{Expansion. }Starting from the root node, the generator traverses the grammar tree to construct a derivation tree, which ultimately forms a complete prompt. During traversal, if the current node is an AND node, all child nodes are recursively visited. For other node types, the process enters the selection phase.

\textbf{Selection.} If the current node is a RAND node, the number of child node visits is determined randomly within a predefined range; If the current node is an OR node and has unvisited children, one is selected at random; If all children have been visited, the Upper Confidence Bound formula is used to compute weights for each child, and a child node is selected via weighted random sampling.

\begin{equation}
\begin{split}
C(v) :=\ & \operatorname{weighted}_{v' \in \operatorname{Children}(v)} \left( Q(v, v') + \sqrt{\frac{2 \ln N(v)}{N(v, v')}} \right)
\end{split}
\end{equation}


\begin{table}[t]
\caption{The number of images generated by the wanx2.0-t2i-turbo model that bypassed the detector (first 200 rounds).}
\label{score-table}
\vskip 0.15in
\begin{center}
\begin{small}
\begin{sc}
\begin{tabular}{l r}
\toprule
Round  & Count \\
\midrule
0--49    & 0 \\
50--99   & 3 \\
100--149 & 1 \\
150--199 & 1 \\
\bottomrule
\end{tabular}
\end{sc}
\end{small}
\end{center}
\vskip -0.1in
\end{table}

Let $Q(v, v')$ be the probability of generating successfully forwarded requests by the CDN after selecting sub-node $v'$ under node $v$, where $N(v)$ denotes the number of times that node $v$ has been visited, and $N(v, v')$ represents the number of times that sub-node $v'$ has been selected under node $v$. In this context, $Score_i$ represents the probability that the detector classifies an image as AI-generated, where a value of 1 indicates a 100\% probability of being identified as AI-generated.


\begin{equation}
Q(v, v') = \frac{1}{N(v, v')} \sum_{i=1}^{N(v, v')} 2 \times (1 - \text{score}_i)
\end{equation}
\textbf{Simulation Phase}: The prompt generator combines the visited leaf nodes into a prompt, which is then passed to the T2I model to generate an image. The AIGC detector will check the generated image and output an AI-generated confidence score, $Score_i$.

\textbf{Back propagation Phase}: Based on the score given by the detector, the generation parameters\( Q(v, v') \) and \( N(v, v') \) associated with the visited nodes in the grammar tree are updated accordingly..




\section{Evaluation}
\subsection{AIGC adversarial Detection Competition}
We participated in an AIGC adversarial Detection Competition~\cite{tencent_zhuque_2025} organized by Tencent,
The organizer provided Tencent's self-developed AIGC detector, Zhuque~\cite{zhuquedetector}. 

Participants were tasked with designing prompts to generate portrait images. The objective was to design prompts to generate images that bypass the Zhuque AIGC detector. A bypass was considered successful if the generated samples achieved an AI detection rate below 50\%. Participants were required to use a limited number of AIGC series APIs, such as Stable Diffusion~\cite{esser2024scaling}, Flux~\cite{flux2024}, and Qwen~\cite{wang2024qwen2}, for image generation. Crucially, generated images had to exhibit normal physiological structures, adhere to physical common sense, and be logically consistent with reality. Furthermore, participants were strictly required to generate images directly using the chosen model, without employing any secondary editing tools, including built-in photo enhancement, removal, or watermark tools.

We utilized our tool with the wanx2.0-t2i-turbo~\cite{aliyun_text2image_v2_en} T2I model API to generate a sufficient number of images capable of bypassing the detectors. This demonstrated that our tool can automatically generate a large quantity of compliant images that successfully bypass the Zhuque AIGC detector, thereby showcasing its effectiveness in real-world scenarios.

\subsection{Testing on Open-source Detector}

To demonstrate our tool's efficacy with other T2I models, we conducted experiments on both the wanx2.0-t2i-turbo~\cite{aliyun_text2image_v2_en} and flux-dev models, using PatchCraft~\cite{zhong2024patchcraftexploringtexturepatch} as the detector.


\begin{table}[t]
\caption{The number of images generated by the flux-dev model that bypassed the detector (first 400 rounds).}
\label{score-table2}
\vskip 0.15in
\begin{center}
\begin{small}
\begin{sc}
\begin{tabular}{l r}
\toprule
Round       & Count \\
\midrule
0--99       & 51 \\
100--199    & 56 \\
200--299    & 59 \\
300--399    & 59 \\
\bottomrule
\end{tabular}
\end{sc}
\end{small}
\end{center}
\vskip -0.1in
\end{table}

We made a particularly interesting discovery: with the PatchCraft detector, most images generated by models we considered more advanced (both intuitively and by their parameter count) were classified as AI-generated by the detector, the results show in~\cref{score-table}. In contrast, weaker models like flux-dev were able to produce a significant number of images that bypassed the detector even within the first 100 iterations.

We analyzed this phenomenon and hypothesize that it's due to the PatchCraft detector's limited training data. We believe PatchCraft's heavy reliance on traditional GAN-generated datasets during training has resulted in insufficient generalization capability for detecting images from newer T2I models. Moreover, most of the training datasets for the model are in 256×256 resolution, resulting in unstable detection performance on high-resolution images.

As~\cref{score-table2} illustrates, our prompt generator can consistently produce images that bypass detectors. Furthermore, as the number of rounds increases, the proportion of prompts generated that achieve the desired target also rises.

\section{Conclusion}
In this paper, we present a novel adversarial prompt generation framework for T2I synthesis that effectively deceives state-of-the-art AIGC detectors. Addressing the challenges of limited controllability and the inefficiency of manual prompt engineering, we introduce a grammar tree-based generator combined with the UCT-Rand algorithm to explore the prompt space efficiently. Our method automates the creation of diverse, semantically rich prompts and significantly improves evasion success across multiple generative models. Extensive experiments show strong attack performance against both open-source and commercial detectors, revealing vulnerabilities to semantic-level prompt perturbations. In addition, our tool can be used to generate more diverse datasets for robustness training of AIGC detectors, and to support the improved design of AIGC detection frameworks.

\bibliography{main}
\bibliographystyle{icml2025}

\end{document}